# IMPROVING FEW-SHOT IMAGE CLASSIFICATION THROUGH MULTIPLE CHOICE QUESTIONS

Emmett D. Goodman, Dipika Khullar, Negin Sokhandan, Sujitha Martin, Yash Shah

Generative AI Innovation Center
edmgood@, dikhulla@, ngnsl@, sujimart@, syash@amazon.com

## ABSTRACT

Visual Question Answering (VQA) models have shown an impressive potential in allowing humans to learn about images using natural language. One promising application of such models is for image classification. Through a simple multiple choice language prompt (i.e. "Question: Is this A) a cat or B) a dog. Answer: ") a VQA model can operate as a zero-shot image classifier, producing a classification label (i.e. "B) a dog."). Compared to typical image encoders, VQA models offer an advantage: VQA-produced image embeddings can be infused with the most relevant visual information through tailored language prompts. Nevertheless, for most tasks, zero-shot VQA performance is lacking, either because of unfamiliar category names, or dissimilar pre-training data and test data distributions. We propose a simple method to boost VQA performance for image classification using only a handful of labeled examples and a multiple-choice question. This few-shot method is training-free and maintains the dynamic and flexible advantages of the VQA model. Rather than relying on the final language output, our approach uses multiple-choice questions to extract prompt-specific latent representations, which are enriched with relevant visual information. These representations are combined to create a final overall image embedding, which is decoded via reference to latent class prototypes constructed from the few labeled examples. We demonstrate this method outperforms both pure visual encoders and zero-shot VQA baselines to achieve impressive performance on common few-shot tasks including MiniImageNet, Caltech-UCSD Birds, and CIFAR-100. Finally, we show our approach does particularly well in settings with numerous diverse visual attributes such as the fabric, article-style, texture, and view ,of different articles of clothing, where other few-shot approaches struggle, as we can tailor our image representations only on the semantic features of interest. Code will be made publicly available.

## KEYWORDS

Visual Question Answering, Few-Shot Classification, Prompt Engineering

## 1. INTRODUCTION

Image classification, a fundamental task in computer vision, continues to play a transformative role across industries ranging from healthcare [1, 2] to agricultural [3, 4]. Originally consisting of CNN architectures [5, 6] trained on million-scale image datasets [7], many image classification systems can reach, and often surpass, human-level performance [1, 8, 6]. Nowadays, when one needs to develop an image classification model for a new task, the go-to approach is to take a state-of-the-art model, pretrained on ImageNet (or other large-scale image datasets), and fine-tune the classifier on the new dataset [9]. Typically, the required dataset may need to contain over 100 images per class, and the model needs to be re-trained and optimized with new hyperparameters and procedures specific to the task at hand. While this method is broadly applicable across almost any classification task, it requires both significant data resources per class, and careful model training and validation. For many applications, it may not be possible to rapidly acquire such a relevant dataset. In any application, retraining a model can be time-consuming, as learning rate, data augmentation strategies, and other hyperparameters may be dataset and task specific. Furthermore, in dynamic systems, if a new class is added, the model needs to be retrained - not ideal for online production settings where the dataset may change with time.

Training-free, data-efficient approaches for image classification are key for rapid and robust model development in dynamic production environments.

At the limit of having no training data, multi-modal visual-language models have emerged as appealing approaches for zero-shot image classification [10, 11]. Such approaches side-step both requirements for large datasets, and custom training procedures. The most prevalent approaches, such as those inspired by CLIP (Contrastive Language-Image Pre-training [11])-based architectures, comprise an image encoder and a language encoder, mapped to the same latent space, to provide a zero-shot image classification label. Since 2021, hundreds of papers inspired by CLIP have emerged, targeting various forms of zero- and few-show image understanding tasks [12].

With the recent rise of high-powered large language models (LLMs) [13], there are emerging opportunities to take advantage of the inductive bias present in LLMs to further bolster the performance of multi-modal models for data-efficient image classification. Visual Question Answering (VQA) models, comprised of an image encoder, language encoder, and language decoder, the latter two sometimes paired in the form of an LLM, are an appealing way to incorporate LLM advances to continuously improve image classification performance in a data-efficient way. These VQA models ingest an image and a question, and output a language answer, and therefore can directly act as zero-shot image classifiers. In this work, we explore the possibility and performance of using BLIP-2 (Bootstrapping Language-Image Pre-Training [14]), with carefully-chosen multiple choice question prompts and latent representations, to directly perform as a zero-shot image classifier.

For specialized classification tasks, image data may be quite domain specific, and language labels may be unfamiliar to a VQA model. In these cases, just a handful of examples could, in theory, significantly improve model performance. However, without full end-to-end training (which can be quite challenging with less than 5 examples) it is unclear how to incorporate this limited training set. We demonstrate a simple way to transition VQA models from the zero-shot to the few-shot setting, to significantly improve VQA performance models for image classification. Using the VQA model as a language-conditioned image encoder, we can compare various extracted latent representations to representations from a handful of labeled examples. We show that by extracting the right set of activations, this simple, training-free approach, outperforms both vision-only image classifiers, as well as zero-shot VQA models, by a significant margin.

## 2. RELATED WORK

### 2.1. Zero-Shot Image Classification

Zero-shot classification is the machine learning problem of identifying image classes which were not directly seen in a training phase. One of the recent large-scale successes in zero-shot image classification was CLIP (Contrastive Language-Image Pre-training [11]), where sets of image and their corresponding language captions were embedded into a shared latent space. By matching an image with multiple candidate language queries in this latent space, CLIP provides class probabilities, even for textual labels before unseen to the model. Since then, stepwise improvements have largely targeted improved model structure, increased dataset sizes, and modified loss functions, but with the same high-level multi-modal contrastive objective [12].

A recent example takes this a step further by using large language models. Rather than modifying CLIP itself, which matches an image embedding with an embedding of the textual class label, each class label is first fed to GPT-3 to generate a detailed language description, which is then fed to the CLIP language encoder [15]. In another example motivated by generative models, this time in vision, Udandarao et al. utilize stable diffusion to generate a support set of synthetic images, from well-defined language prompts, which were used to more effectively align the image-text latent space for improved zero-shot performance [16].

### 2.2. Few-Show Image Classification

Few-shot image classification extends the zero-shot task by providing only a few labeled training examples, typically no more than five per class. Compared to zero-shot image classification, which typically has a language component, few-shot image classification typically would not. For example, a historically common approach to this task involves meta-learning - or learning how to learn [17, 18]. A meta-training set is broken into a set of smaller training and validation sets, each representing a few-shot problem. On this meta-training set, an optimizer, network, or distance metric is developed to perform well in the few-shot setting. Once developed, the strategy is evaluated on a meta-test set, consisting of new few-shot problems. These methods may involve further model tuning at test time, ranging from the entire neural network, to a simple linear classifier for each few-shot problem. However additional work suggests that rather than focusing on meta-learning algorithms, which are often complex, focusing on better embeddings, or representation learning methods, may be a more effective way forward in few-shot image classification [19]. The notion of better embeddings is one targeted in our present work.

Our approach can be described as multimodal few-shot image classification; here we are using language information, provided by the LLM, to improve model performance. A few other examples of multimodal approaches for few-shot image classification exist in the literature, and these combine information, including text or speech, to improve performance [20]. Schwartz et al. show how language can be used to iteratively refine vision-only latent prototypes [21]. Here, a vision prototype for each image class is first created via a visual encoder, and this prototype is iterative refined using language branches via task-category semantics.

### 2.3. Visual Question Answering Models

Visual Question Answering (VQA) models are multi-modal models that take an input image and a language question, and output a language response. Commonly, VQA models are tested under zero-shot conditions, such as on the zero-shot VQAv2 task [22, 23]. The zero-shot regime is particularly relevant for VQA tasks, due to the large number of possible responses for any language question across complex images.

Although earlier VQA models only used a language decoder [14, 24] recent, state-of-the-art VQA models have begun integrating entire pretrained LLMs into their structure. As an example, PaLI [25] leverages separate large unimodal backbones for language and vision modeling to transfer capabilities and reduce training costs. For many of these VQA models, there is no need to even fine-tune the LLM, which can be used with their original weights [14, 26]. BLIP-2 [14] is a performant recent example of this, where the authors propose a new approach called a Querying Transformer (Q-Former) for achieving effective vision-language alignment with frozen unimodal models that had been pre-trained using multiple different vision-language objectives. The Q-Former is a lightweight transformer that utilizes learnable query vectors to extract visual features from a frozen image encoder. It acts as an information bottleneck between the image encoder and the frozen LLM.

Another example of this class of VQA architectures is MiniGPT-4 [26] which incorporates the Vicuna language model (LLM) as its language decoder, which achieves a performance approaching that of ChatGPT. For visual perception, MiniGPT-4 employs the pretrained vision component of BLIP-2. Due to ease of use, our work focuses on using BLIP-2 as a VQA model, but techniques should translate to new VQA models as they are developed.

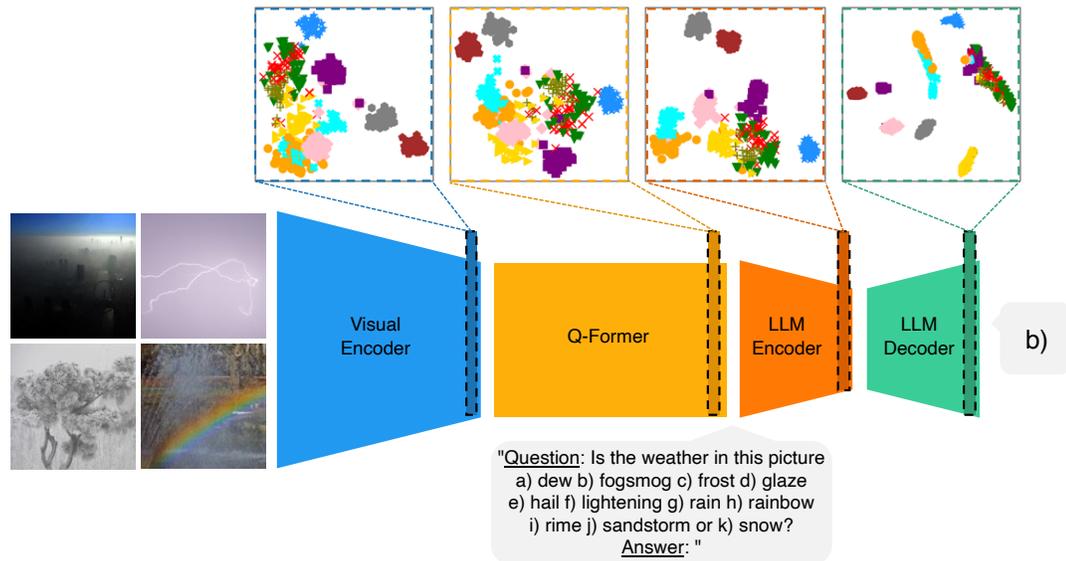

**Figure 1: Proposed method for few-shot image classification using a VQA model.** The VQA model, BLIP-2, acts as a language-steerable image encoder and representation extractor. Image representations are extracted from four different locations in the network, and are assigned a class label via nearest-neighbor lookup in the latent space. In the inset plots, each color represents a distinct image class.

## 3. METHODS

### 3.1. Approach

Our core task is multi-class classification under the following constraints: (1) there are only have few (i.e. 1-5) image labels per class and (2) the category text label is available. We tackle the few-shot image classification task using latent representations from a VQA model.

VQA models are appealing because they can be used as zero-shot image classifiers, in scenarios where no data is available, due to their vision-language pre-training. There are clear reasons and incentives to investigate VQA models in the few-shot regime, where some image-label pairs are available as domain-specific examples. While generally performant, on any new domain-specific task, zero-shot VQA model performance likely will not be as performant as a model specifically trained on data for that specific task. In theory, VQA models could be retrained for each task, but in sparse data conditions (1-5 examples per class) this can be extremely challenging due to overfitting. Aside from end-to-end training, it's not obvious how to best integrate small amounts of domain knowledge into these zero-shot models.

We present a method that can build on the zero-shot performance of VQA models by introducing the model to just a few examples, without any model training. Shown in **Figure 1**, rather than training the VQA model, we simply use it to retrieve various image embeddings. These latent representations are created by querying the image with a multiple-choice question. This produces a series of intermediate multimodal representations which may be useful for the downstream classification task. Past research has indicated that a good embedding may be all you need for good few-shot performance [19]; specifically, we demonstrate that a very specific slice of activations coming out of the LLM-decoder provide highly-informative representations for few-shot image classification. In many cases, this LLM-decoder representation can be further enriched by concatenating a purely visual representation produced by the vision encoder.

### 3.2. Datasets

*Black-and-White Pets*: This is a toy dataset used to explore the tunability of latent representations through language. It comprises two image classes - black-and-white (*BW*) pets, and full-color (*RGB*) pets. The pets include images of cats and dogs, and each pet image in the RGB class has a corresponding decolorized image version in the BW class. Dog images were taken from the Stanford Dogs Dataset [27], and cat images were taken from the Kaggle Cat Dataset [28].

*MiniImageNet Dataset* [29]: The MiniImageNet dataset is a subset of the ImageNet dataset, specifically designed for few-shot learning tasks. It consists of 100 classes with 600 images per class, for a total of 60,000 images. Each image has a resolution of 84x84 pixels and is categorized into one of the 100 fine-grained classes.

*CUB200 Dataset* [30]: The CUB200 (Caltech-UCSD Birds-200-2011) dataset is focused on fine-grained bird species classification. It consists of 200 bird classes, with a total of 11,788 images. The images in CUB200 are higher-resolution compared to the previous datasets, with varying dimensions. The images were resized (maintaining aspect ratio) and center cropped to 64x64, 128x128 and 256x256 resolutions.

*CIFAR-100 Dataset* [31]: The CIFAR-100 dataset is based on the CIFAR-10 dataset and is more challenging due to the larger number of classes and greater class diversity. It contains 100 classes, with each class having 600 images. Each image has a resolution of 32x32 pixels.

*DeepFashion MultiModal Dataset* [32]: The DeepFashion MultiModal dataset is a fashion dataset containing a vast collection of images and textual information capturing clothing items, styles, and contexts. In particular, this dataset spans two genders (female, male), eight fabrics, seven patterns, and five viewpoints. The resolution of the images in the DeepFashion MultiModal dataset can vary, but typically they are in the range of a few hundred pixels on each side.

### 3.3 Implementation Details

Each classification problem studied consists of a dataset, comprised of n-distinct classes, with 20-50 image examples per class. For MiniImageNet and CIFAR-100, the problem was posed as a 100-class classification problem, with 20-to-50 instances from each class present. For the CUB dataset, we used 50 different classes, with 20 instances per class. This is a distinct and significantly more challenging scenario than the common 5-way few-shot classification problem, which only studies 5 classes at a time.

Each dataset has an associated language prompt, which unless mentioned otherwise, is formulated in the form of "*Question: Is this a picture of a) __, b)__, ...? Answer:* ". This prompt form is motivated by the prompts used in the BLIP-2 paper [14]. A VQA model from BLIP-2 was used, comprised of a ViT-L/14 vision encoder, a Flan-T5-XL LLM and a Querying Transformer (Q-Former) to bridge the modality gap between image encoder and LLM.

As a baseline, zero-shot performance (ZSP) is evaluated by comparing the VQA language output to the original multiple-choice letter presented in the language prompt. In the majority of cases, the VQA directly outputs the multiple-choice answer, such as "a". When more than a single letter is produced by the model (such as in cases where the model outputs "a)"), the first character is taken as the zero-shot answer.

For few-shot performance, each class is separated into a "train" and "test" set, where the train set contains between 1-5 images per class, and the test set contains the remaining images. Both the train and test data are passed to the VQA model, which produces various latent representations for each image. Specifically, for each image, four representations were extracted: a representation from the visual encoder (torch.size[257,1408]), a Qformer representation (torch.size[1,32,768]), an LLM-encoder representation (torch.size([64,2048])), and an LLM-decoder representation (torch.size([5,n,2048])), in addition to the zero-shot language output **(Figure 1)**. These were extracted by adding model hooks at *model.visual_encoder*, *model.Qformer.bert*, *model.t5_model.encoder*, *model.t5_model.decoder* respectively. Each of these four representations was averaged across all dimensions except the final dimension, to produce embeddings of size torch.size[1,1408], torch.size[1,768], torch.size[1,2048], and torch.size[1,2048]. The shape of LLM decoder activations varied per sample and dataset (i.e. torch.size([5,n,2048]) and taking the torch.size([5,1, 2048] activation led to the best performance for this task. This will be studied in a simple ablation study below. A final embedding, which performed well across tasks, was created by concatenating the LLM-decoder embedding and visual-encoder embedding, to create a multi-modal embedding of shape torch.size[1,3456].

For each type of embedding, the unlabeled test data are compared, in Euclidean distance, to the latent representations of the labeled training data, and are assigned class labels based on which labeled latent representation is closest. This can be interpreted as K-Nearest Neighbors (KNN) in the latent space. We explored the effect of the "K" hyperparameter, and due to the limited number of training samples, found K=1 to generally be the best. This is therefore equivalent to nearest neighbor lookup. Furthermore, we observe that averaging the few-shot examples in the latent space provides better overall performance across diverse classification tasks, so this was performed for each experiment. This is similar to defining a prototype embedding for each class, and identifying the prototype each test datapoint is closest to. Prior to nearest-neighbor lookup, each vector was normalized by subtracting the mean of the test embeddings and dividing by the standard deviation of the test embeddings. For each set of conditions, the experiment is repeated 50 times, due to the variable nature of few-shot classification (and its heavy dependence on the combination of labeled examples selected).

## 4. RESULTS AND EVALUATION

### 4.1. Toy Example: Black-and-White Pets

We start with a toy example to motivate the utility of multi-modal embeddings for few-shot image classification. This dataset, we call the *Black-and-White Pets dataset*, is comprised of two image classes - black-and-white (*BW*) pets, and full-color (*RGB*) pets. The task is to predict whether an image is *BW* or *RGB*. Both classes are comprised of cats and dogs.

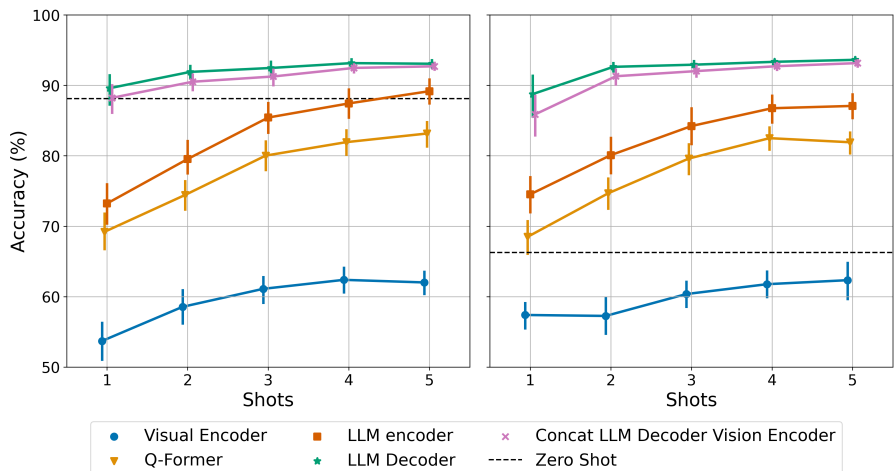

**Figure 2. Model performance as a function of embedding type, number of labeled training images, and prompt.** Left prompt: Question: Is this a picture of a) a black-and-white pet or b) an in-color pet? Right prompt: Question: Is this a picture of a) a black-and-white pet or b) an RGB pet?

We first investigate which VQA-generated representations are best for this binary classification task. Results are shown in **Figure 2** for two different multiple-choice questions.

Across all embedding types, we observe clear improvements in performance as we increase the number of labeled datapoints per class. This can be attributed to the fact that averaging more embeddings creates a more accurate prototype embedding to refer test data to, by averaging out embedding components that are not directly related to key class attributes.

Interestingly, for both multiple-choice questions, we find the least effective representations are those generated directly from the ViT visual encoder, which achieves ~55% 1-shot accuracy, and ~62% 5-shot accuracy, independent of language prompt. Then, in order from least performant to most performant (**Figure 2, left**), we have Q-Former embeddings (69.20% 1-shot), LLM-encoder embeddings (73.23% 1-shot), and finally LLM-decoder embeddings (89.59% 1-shot). This order corresponds from least to most language mixing in the latent representation, as we move deeper into the VQA model. Finally, we have the concatenated language/vision embeddings, performing similarly to the LLM-decoder embeddings, at (88.23%).

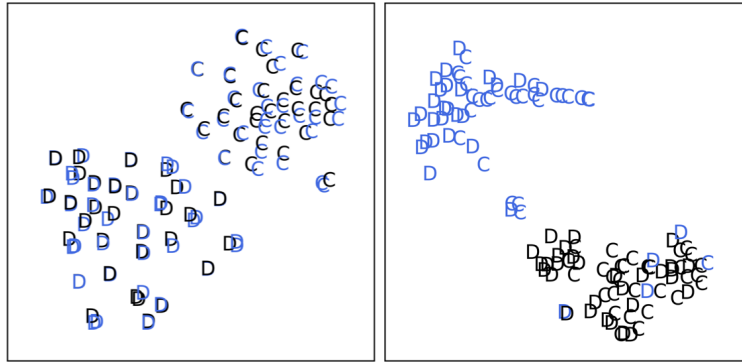

**Figure 3. t-SNE plot of Black-and-White Pets embeddings at different stages in the captioning model.** Embeddings (left) directly after ViT vision encoder, and (right) after the LLM-decoder. C=cat, D=dog; black color represents *BW* images, blue represents RGB images.

The embeddings from the Q-Former, while prompt-agnostic and pre-prompt injection, reformulate the image embedding in terms of LLM language embeddings, and may capture some elements of image color details, perhaps explaining better performance than those directly after the ViT image encoder.

While visual and Q-Former embeddings are not impacted by the prompt language, we do observe the multiple-choice prompt impacting downstream LLM-encoder and LLM-decoder embeddings. The multiple-choice question language prompts steer the representations produced by the LLM-encoder and LLM-decoder. To demonstrate, we select two different VQA prompts and explore the impact on model performance. In **Figure 2, left** (better prompt) we observe high performance for the zero-shot baseline (88.75% 1-shot), while in **Figure 2, bottom** (worse prompt) we observe mediocre performance (62.25% accuracy). While it may not be clear *a priori* which prompt would be better for this task, we do highlight classification sensitivity to the exact question phrasing as a weakness of zero-shot image classification. Interestingly, with the changed language prompt, we observe the 1-shot classification accuracy for the LLM-decoder embeddings to drop from 88.24% to 86.51%, or about 2%. While this is notable, it's much less significant than the nearly ~26% drop of the zero-shot performance. We find that the LLM-produced representations are sensitive to the prompt, but still robust enough to maintain good performance. Importantly, we observe the proposed approach exceeds zero-shot performance by significant margins (between 7% and 36% depending on the prompt).

Interestingly, given an adversarial prompt (i.e. Q: Is this an a) apple or b) orange), we observe severely affected performance (1-shot LLM-decoder accuracy decreases from 88.24% to 62.33%, figure not included). This further emphasizes that these representations are controlled by the natural language question.

We note that the BLIP-2 visual encoder is alone actually quite performant for other tasks (i.e. **Table 1**). Nevertheless, we observe that for the Black-and-White pets dataset and task, image embeddings are very poor for classification - approaching that of a random guess (i.e. 50%). This can be visualized in the t-SNE plot in **Figure 3, left**. We observe that the vision encoder clearly separates cats from dogs, but completely fails to separate *BW* from *RGB*, the targeted task. This is likely because most vision encoders are trained on tasks related to the semantics of the object in the image (i.e. cat vs dog) and not on other style aspects of the image (i.e. *RGB* vs *BW*). However, as we move deeper into the VQA network, the impact of the language query takes increasing effect, and can be observed in **Figure 3, right**, where we observe effective separation of *BW* from *RGB* images. Using words, we convey to the model that we care about those color elements, and can now observe clear separation between BW and RGB images.

## 4.2. Few-Shot Benchmarks

| Dataset | Num Classes | Zero-Shot Performance | One-Shot Performance | | | Five-Shot Performance | | |
|---|---|---|---|---|---|---|---|---|
| | | Language Output | Visual Encoder | LLM-Decoder | Concat Vis/LLM | Visual Encoder | LLM-Decoder | Concat Vis/LLM |
| MiniImageNet | 100 | 49.86% | 62.99% | 73.14% | **75.96%** | 85.84% | 83.79% | **87.72%** |
| CUB50 | 50 | 6.69% | 41.33% | 39.78% | **46.69%** | 67.33% | 54.02% | **68.70%** |
| CIFAR-100 | 100 | 48.70% | 57.06% | 60.43% | **64.25%** | 79.99% | 74.13% | 78.83% |

Table 1. Few-shot performance on benchmark tasks as a function of labeled examples and selected representation.

To highlight the performance of the approach, we take aim at classic few-shot image classification benchmarks, including MiniImageNet, CUB, and CIFAR-100 (**Table 1**). As ImageNet is a prototypical task for image classification, and most image encoders are first pre-trained on ImageNet, we thought this may be a particular case where additional language guidance would have little to add. However, we observe that for MiniImageNet, our LLM-decoder embeddings significantly outperform visual embeddings by ~10% in the one-shot regime. Furthermore, when we concatenate the visual embeddings with the LLM-decoder embeddings, we achieve further gains: an additional ~3% accuracy. We conclude that even for vision-centric tasks, language-guided representations are useful.

Caltech-UCSD Birds dataset is a vision task focused on fine-grained classification tasks, and we observe that our proposed approach achieves improved performance over the visual-only baseline in distinguishing between 50 types of birds. We note (not reported in detail here) that in extending from CUB50 to CUB200, Concat Vis/LLM embeddings no longer outperform purely visual embeddings. This is likely due to the fact that there is significant textual overlap in class labels (i.e. "White-crowned sparrow" versus "white-throated sparrow"), that textual class labels lend no additional advantage. This also manifests itself in a zero-shot performance which is <1 % when studying all 200 fine-grained bird classes.

For CIFAR-100, we observe a 3% improvement using LLM-Decoder embeddings over the Visual Encoder embeddings, and this jumps to 7% improvement when we use the Concat Vis/LLM embeddings.

While we observe consistent benefit of language in the one-shot regime, the gap between language-based and vision based few-shot performance narrows with more image data, moving into the 5-shot regime. Nevertheless, we still observe 2% improvement using language on MiniImageNet, while achieving near parity on CUB and CIFAR-100 datasets.

## 4.3. DeepFashion

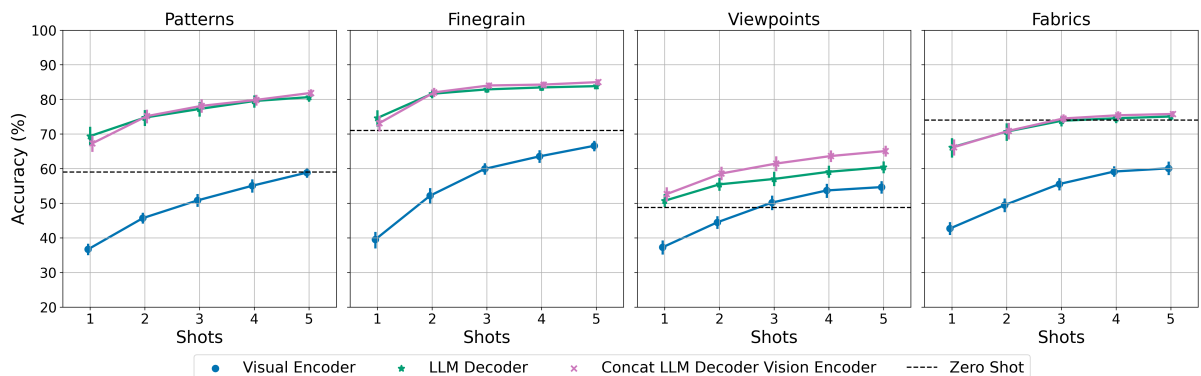

Figure 4. Performance on DeepFashion MultiModal across various slices of the data: patterns, finegrain, viewpoints, and fabrics. **Prompts (left to right):** [Patterns] - Question: Is this a picture of a) floral, b) pure color, c) lattice, d) graphic, e) striped? [Finegrain] - Question: Is this a picture of a) Seamed Skater Dress, b) Abstract Satin Front Top, c) Paisley Pattern Pencil Skirt, d) Shirred Sateen Maxi Skirt, e) Embroidered Floral A Line Dress, f) Open Knit Raglan Sweater, g) Open Knit

Cowl Neck Sweater, h) Rugby Striped Boxy Tee? [Viewpoints] - Question: Is this a picture of a) full view, b) front view, c) back view, d) side view? [Fabrics] - Question: Is this a picture of a) chiffon, b) denim, c) knitted, d) cotton?

The last dataset studied in this work is a dataset of diverse fashion images from the Deepfashion MultiModal dataset [32]. DeepFashion MultiModal is a large-scale garment dataset containing 44,096 high-resolution clothing images. Each image has been annotated with multiple clothing characteristics, including shapes, textures, image viewpoint, and clothing patterns. In our study, DeepFashion is of particular interest because it presents a multi-label classification dataset, where a single image can have multiple orthogonal labels. For each of these label types, we sample subsets of the DeepFashion dataset. Within each subset, we examine different classification problems: pattern, fine-grained article type, viewpoint, and fabric-type classification.

**Figure 4** shows the results of our approach across these four classification problems. We observe that for each problem, regardless of number of shots, the Concat Vis/LLM and LLM Decoder embeddings yield the highest classification accuracy. These two language-infused embeddings outperform purely visual embeddings by a margin ranging from 10% to 32%. In a comparison of language-infused versus vision-only embeddings, the multi-attribute dataset reveals that while Concat Vis/LLM and LLM Decoder embeddings are able to adapt effectively to each new problem through language prompts and demonstrate high performance, the Vision Encoder is consistently less effective, as it is uncertain which aspects of the fashion dataset to focus on.

Notably, out of all four classification problems, our approach performs the best on the "finegrain" problems, which have approximately twice as many classes as the other three problems (i.e. 8 classes). Although initially surprising, this underscores the power of language. The finegrain classification problem offers the most vivid descriptions of object categories, such as "Paisley Pattern Pencil Skirt". In this context, the increased language information can likely more than compensate for the additional classes.

### 4.4. Selection of Latent Representations

Due to how LLMs decode internal representations into a variable-length sequence of tokens, for any given query to the VQA model, the LLM-decoder representation could vary from size between *torch.size([5,5,2048])* to *torch.size([5,30,2048])*. Nevertheless, these representations are ultimately decoded into a simple string such as "a" or "b". To better understand this relationship, we studied the impact of taking a subset of the LLM decoder embeddings, and not necessarily taking the entirety of the produced LLM-decoder activation. We observe that as we sub-select slices in the $2^{nd}$ dimension, the performance of the embedding slice varies dramatically (**Figure 5**). For example, while the $2^{nd}$ embedding position provides an accuracy of 88.23 $\mp$ 8.44%, the 12th embedding achieves a much lower accuracy (61.78 $\mp$ 13.54%), or a ~27% performance drop. Although **Figure 5** is a demonstration on the Black-and-White Pets dataset, similar results were observed on other datasets studied. For this reason, for all manuscript experiments we only take the second token embedding as the final representation extracted from our LLM decoder (i.e. torch.size([5, 1, 2048]). Due to the inclusion of a start token in the decoded representation, the second token, and its embedding, is most related to the final answer output by the zero-shot decoder. We note that inclusion of the entire pooled decoded LLM representation leads to significantly decreased, and non-competitive, performance (not reported). Overall, we interpret this $2^{nd}$ embedding to contain much richer information than the final zero-shot response created by decoding it.

## 5. CONCLUSIONS, LIMITATIONS, AND FUTURE WORK

In this work, we extend zero-shot VQA image classification to the few-shot domain using a handful of labeled images, a multiple-choice question, and careful choice and manipulation of latent representations. We demonstrate that a VQA model can produce useful representations, steered by natural language, for few-shot image classification. We studied the internal representations produced within BLIP-2 VQA model, and show that they can outperform the purely visual image representations, which may pick up on visual elements unrelated to the task at hand. Our approach requires no additional training whatsoever, and should become more powerful as VQA models, backed by pre-trained vision and language foundation models, become more performant.

An important consideration when choosing a modeling strategy, especially in the few-shot regime, is the nature of the classification task. We observe that although our approach offers increased performance under many conditions (i.e. MiniImageNet, CUB, CIFAR-100, weather dataset (not shown)), this approach may not universally be the most performant. Especially when there is not much visual diversity with a class, or the language labels for multiple classes are very similar, visual-only classification may be competitive to multi-modal representations.

Future work should focus on a deeper understanding of the semantic content found in each layer of the VQA model. Given the network structure of BLIP-2, it is clear all collected representations prior to the LLM should be unaffected by the language prompt injected into the VQA model; nevertheless, we observe in some conditions, representations within the QFormer module may perform quite well, outperforming the ViT image encoder. Across experiments, we observe interesting differences when we compare ZSP performance with 1-shot LLM decoder performance; sometimes the 1-shot performance significantly outperforms ZSP performance, and other times the 1-shot performance merely asymptotes to the ZSP performance. Understanding why we have this variability may unlock further gains in few-shot performance.